\def\year{2019}\relax
\documentclass[letterpaper]{article} 
\usepackage{aaai19}  
\usepackage{times}  
\usepackage{helvet}  
\usepackage{courier}  
\usepackage{url}  
\usepackage{graphicx}  
\usepackage{amsmath,amssymb}
\usepackage{booktabs}
\usepackage[bottom]{footmisc}

\newcommand{\tabincell}[2]{\begin{tabular}{@{}#1@{}}#2\end{tabular}}

\begin{document}
%
\title{Learning Fully Dense Neural Networks for Image Semantic Segmentation}
\author{Mingmin Zhen\textsuperscript{1}, Jinglu Wang\textsuperscript{2}, Lei Zhou\textsuperscript{1}, Tian Fang\textsuperscript{3}\thanks{Tian Fang is with Shenzhen Zhuke Innovation Technology since 2017.}, Long Quan\textsuperscript{1}\\
\textsuperscript{1}Hong Kong University of Science and Technology, \textsuperscript{2}Microsoft Research Asia, \textsuperscript{3}Altizure.com \\
\{mzhen, lzhouai, quan\}@cse.ust.hk, Jinglu.Wang@microsoft.com, fangtian@altizure.com}
\maketitle
\begin{abstract}
Semantic segmentation is pixel-wise classification which retains  critical spatial information. The ``feature map reuse" has been commonly adopted in CNN based approaches to take advantage of feature maps in the early layers for the later spatial reconstruction. Along this direction, we go a step further by proposing a  fully dense  neural network with an encoder-decoder structure that we abbreviate as  FDNet. For each stage in the decoder module, feature maps of  all the previous blocks are adaptively aggregated to feedforward as  input. On the one hand,  it reconstructs  the spatial boundaries accurately. On the other hand, it learns more efficiently  with the more efficient  gradient backpropagation. In addition, we  propose the boundary-aware loss function to focus more attention on the pixels near the boundary, which boosts the ``hard examples'' labeling. We have demonstrated the best performance of the FDNet on the two benchmark datasets: PASCAL VOC 2012, NYUDv2  over previous works when not considering training on other datasets. 
\end{abstract}

\section{Introduction}
\begin{figure}[t!]
	\centering
	\includegraphics[width=\linewidth]{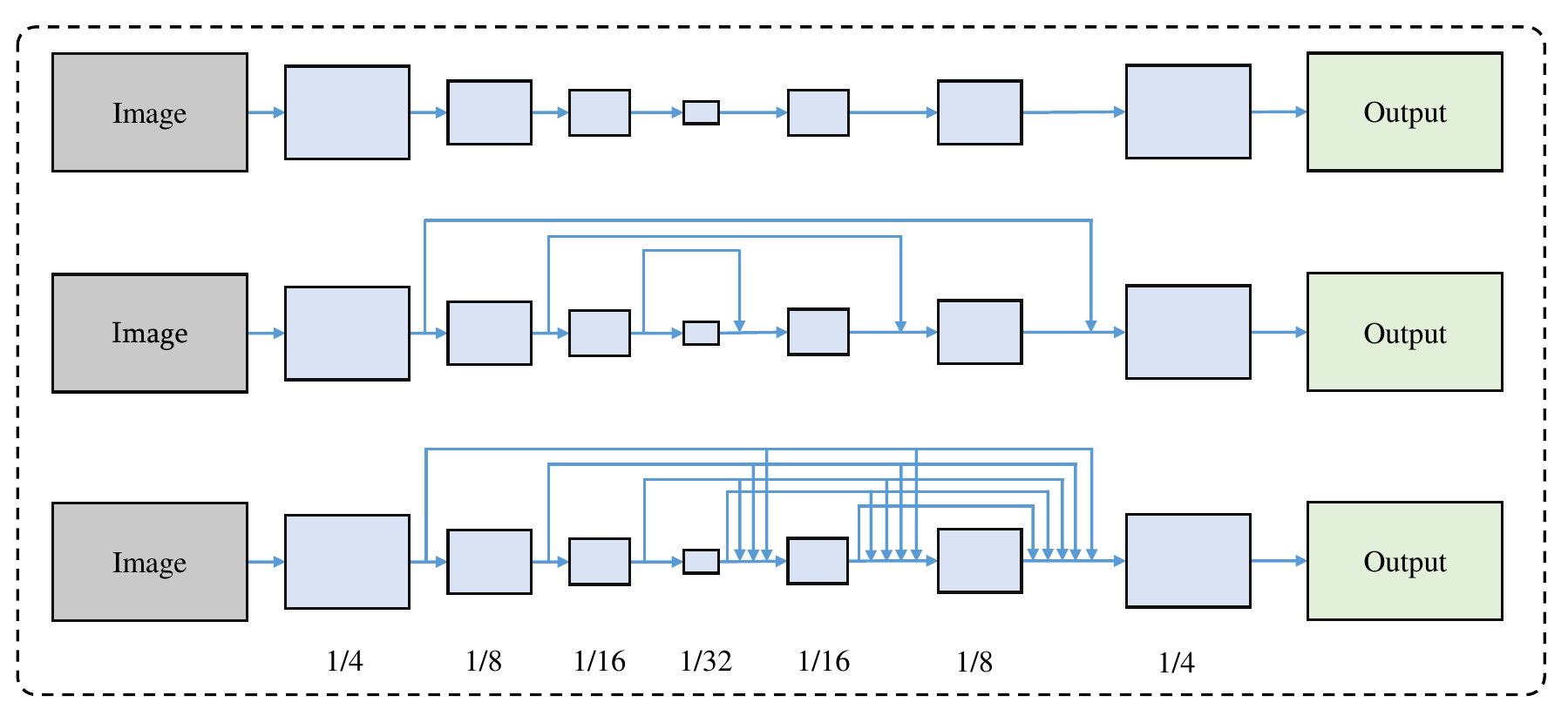}
	\caption{Different types of encoder-decoder structures for semantic segmentation. \textbf{Top}: basic encoder-decoder structure (e.g. DeconvNet \cite{deconvnet} and SegNet \cite{segnet}) using a multiple-stage decoder to predict masks,  often results in very coarse pixel masks since spatial information is largely lost in the encoder module. \textbf{Middle}: Feature map reuses structures using previous feature maps of the encoder module  achieves very good results in semantic segmentation tasks \cite{refinenet,gateFrNet,LRR} and  other related tasks \cite{sharpmask,msdnet}, but the potential of feature map reuse is not  deeply released. \textbf{Bottom}: The proposed fully dense networks, using feature maps from  all the previous blocks, are capable of capturing multi-scale information, of restoring the spatial information, and of benefitting the gradient backpropagation. }
	\label{encoder}
\end{figure}

Recent works on semantic segmentation are mostly based on the fully convolutional network (FCN) \cite{fcn_cvpr}. Generally, a pretrained classification network (such as VGGNet \cite{vggnet}, ResNet \cite{resnet_he} and DenseNet \cite{densenet_huang}) is used as an encoder to generate a series of feature maps with rich  semantic information at  the higher layers. In order to obtain the probability map with the same resolution as the input image size, the decoder is adopted to recover the spatial resolution from the output of the encoder (Fig. \ref{encoder} Top). The encoder-decoder structure is widely used for semantic segmentation \cite{segnet,fcn_cvpr,deconvnet,PSPNet} . \\
\indent The key difficulties for the encoder-decoder structure are twofold. First, as multiple stages of spatial pooling and convolutional strides are used to reduce the final feature map size in the encoder module, much spatial information is lost. This is hard to recover in the decoder module and leads to poor semantic segmentation results, especially for boundary localization.  
Second, 
the encoder-decoder structure is much deeper than the original encoder  network for image classification tasks (such as VGGNet \cite{vggnet}, ResNet \cite{resnet_he} and DenseNet \cite{densenet_huang}). This results in the training optimization problem as introduced in \cite{resnet_he,densenet_huang} though it has been partially solved by using batch normalization (BN) \cite{batchnorm}. \\
\indent In order to address the spatial information loss problem, DeconvNet \cite{deconvnet} uses the unpooling layers  to restore the spatial information by recording the locations of maximum activations during the pooling operation. However, this cannot completely solve the problem since only the location of maximum activations is restored. Another way to deal with this problem is to reuse the feature maps  with rich spatial information of earlier layers. U-Net \cite{u-net} exploits previous feature maps in the decoder module by a ``skip connections'' structure (See Fig. \ref{encoder} Middle). Furthermore, RefineNet \cite{refinenet} refines semantic feature maps from  later layers with fine-grained feature maps from  earlier layers. Similarly, G-FRNet \cite{gateFrNet} adopts multi-stage gate units to make use of previous feature maps progressively. The feature map reuse significantly improves the restoration of spatial information. Meanwhile, it helps to capture multi-scale information from the multi-scale feature maps of earlier layers in the encoder module. In addition, it also boosts information flow and gradient backpropagation as the path from the earlier layers to the loss layer is shortened.   \\
\begin{figure}[t!]
	\begin{minipage}{.6\linewidth}
		\includegraphics[width=0.99\linewidth]{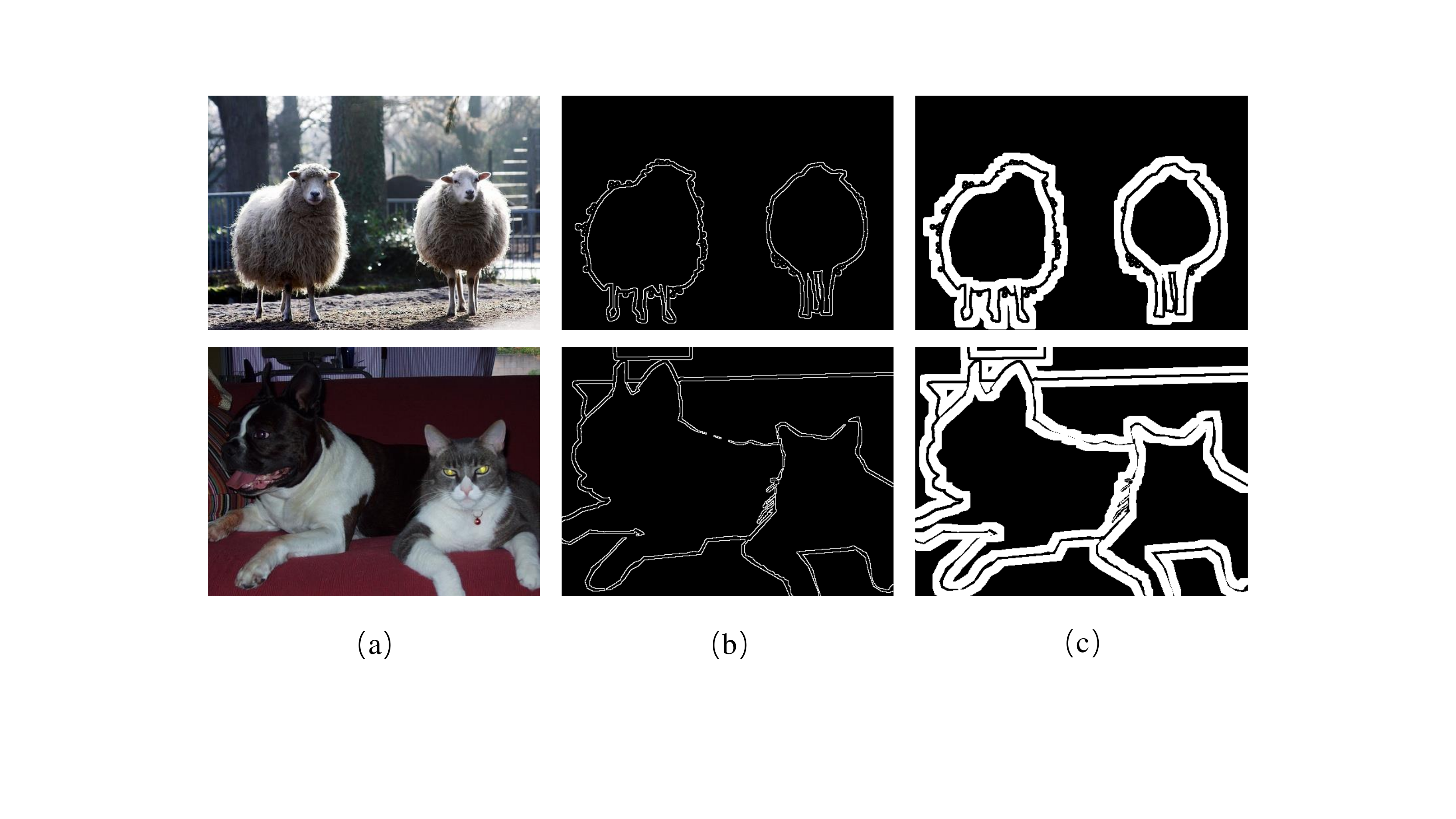}
	\end{minipage}%
	\begin{minipage}{.4\linewidth}
		\includegraphics[width=0.99\linewidth]{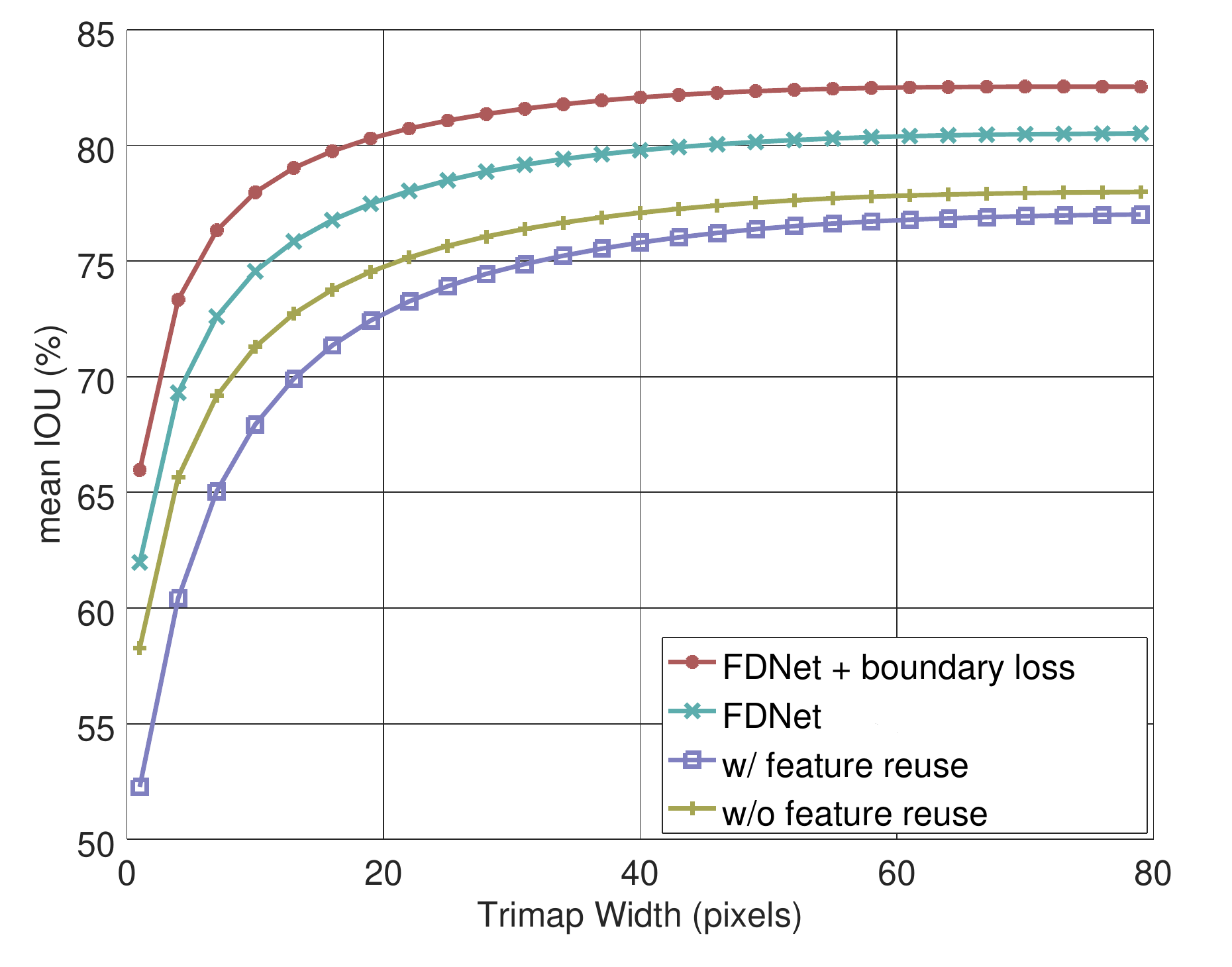}
	\end{minipage}
	\caption{Left: (a) original images; (b) trimap example with  1 pixels; (c) trimap example with  10 pixels. Right: semantic segmentation results within a band around the	object boundaries for different methods (mean IOU).}
	\label{boundary}
\end{figure}
\indent However,  the potential of feature map reuse is not completely revealed. In order to further improve the performance, we propose to reconstruct the encoder-decoder neural network to form a fully dense  neural network (See Fig. \ref{encoder} Bottom). We refer to our neural network as \textbf{FDNet}. FDNet is a nearly symmetric encoder-decoder network and is easy to optimize. We choose  DenseNet-264 \cite{densenet_huang} as the  encoder, which achieves  state-of-the-art results in the image classification tasks. The feature maps in the encoder module are beneficial to the decoder module. The decoder module is operated as an upsampling process to recover the spatial resolution, aiming for accurate boundary localization. The feature maps of different scale sizes (including feature maps in the decoder module) will be fully reused through an adaptive aggregation structure, which will generate a fully dense connected structure. \\
\indent In general, the cross entropy loss function is used to propagate the loss in previous works \cite{parsenet,refinenet}. The weakness of this method is that it sees all pixels as the same. As shown in Fig. \ref{boundary}, labeling for the pixels near the boundary ($\text{band width} < 40$) is not very accurate. In other words, the pixels near the boundary are ``hard examples'', which need to be treated differently. Based on this observation, we propose a boundary-aware loss function, which pays more attention to the pixels near the boundary. Though attention based loss has been adopted in object detection task \cite{focal_loss}, our boundary-aware loss comes from the prior that pixels near the boundary are ``hard examples''. This is very different from focal loss, which pays more attention to the pixels with higher loss. In order to further boost training optimization, we use multiple losses for the output feature maps of the decoder module. As a result, basically each layer of FDNet has direct access to the gradients from the loss layers. This will be very helpful to gradient propagation \cite{densenet_huang}.\\
\section{Related work}
The fully convolutional network (FCN) \cite{fcn_cvpr} has improved the performance  of semantic segmentation significantly. In the FCN architecture,  a fully convolutional structure and bilinear interpolation are used  to realize pixel-wise prediction, which results in coarse boundaries as large amounts of  spatial information have been lost. Following the FCN method, many works \cite{segnet,refinenet,PSPNet} have tried to further improve the performance of  semantic segmentation.\\
\noindent \textbf{Encoder-decoder}. The encoder-decoder structure with a multi-stage decoder gradually recovers sharp object boundaries. DeconvNet \cite{deconvnet} and SegNet \cite{segnet} employ symmetric encoder-decoder structures to restore spatial resolution by using unpooling layers.  RefineNet \cite{refinenet} and G-FRNet \cite{gateFrNet} also adopt a multi-stage decoder with feature map reuse in each stage of the decoder module. In LRR \cite{LRR}, a multiplicative gating method is used to refine the feature map of each stage and a Laplacian reconstruction pyramid is used to fuse predictions.
Moreover, \cite{sdn} stacks many encoder-decoder architectures to capture multi-scale information. Following these works, we also use an encoder-decoder structure to generate pixel-wise prediction label maps. \\
\noindent \textbf{Feature map reuse}. The feature maps in the higher layers tend to be invariant to translation and illumination. This invariance is crucial for specific  tasks such as image classification, but is not ideal for  semantic segmentation which requires precise spatial information, since important spatial relationships have been lost. Thus, the reuse of feature maps with rich spatial information of the previous layers  can boost the spatial structure reconstruction process. Furthermore, feature map reuse has also been used in object detection tasks \cite{fpn} and instance segmentation tasks \cite{sharpmask,mask-rcnn}  to capture multi-scale information when considering the objects with different scales. In our architecture, we  fully aggregate previous feature maps in the decoder module, which shows outstanding performances in the experiments.
\begin{figure*}[t!]
	\centering
	\includegraphics[width=\textwidth]{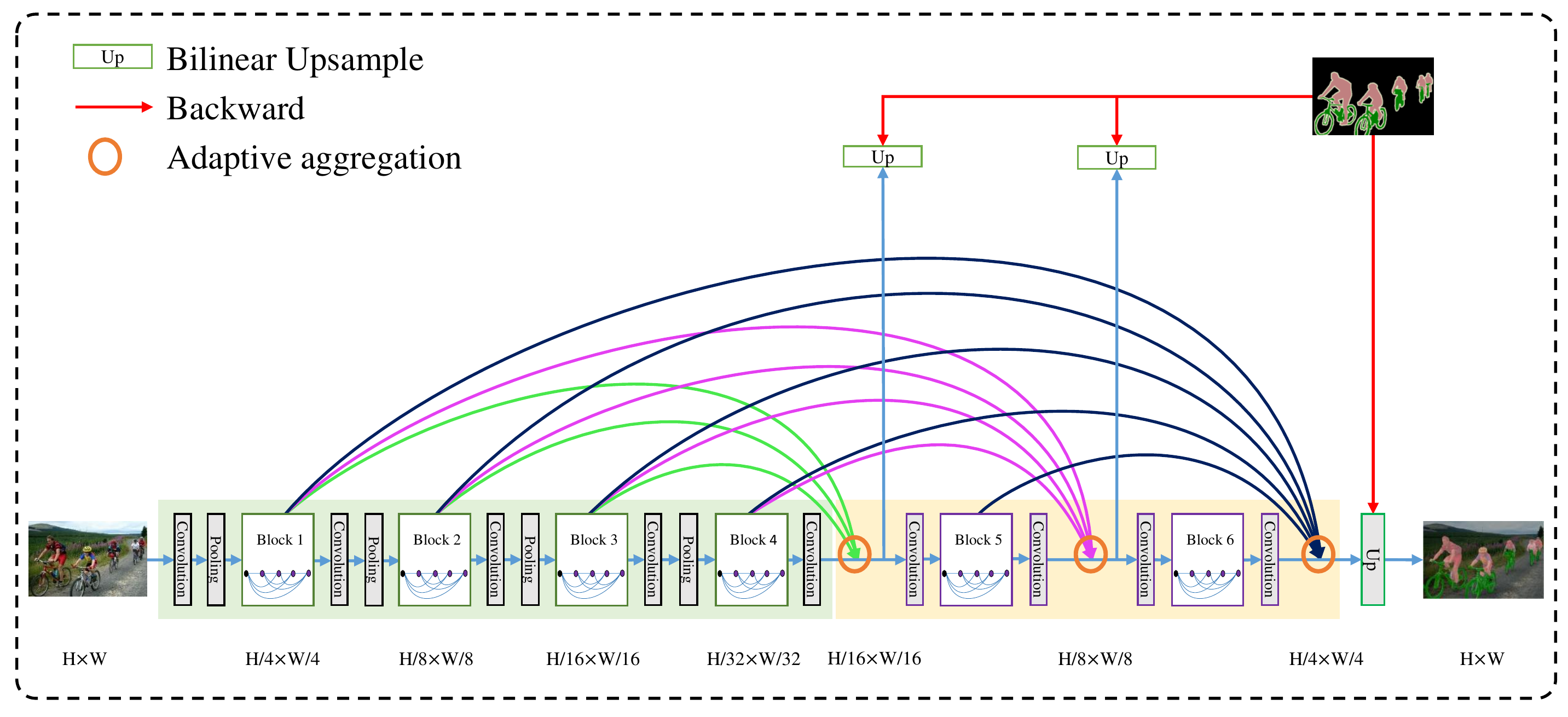}
	\caption{Overview of the proposed fully dense neural network (FDNet). The feature maps (output of dense block $1, 2, 3, 4$) of the encoder module and even the feature maps (output of dense block 5) of the decoder module  are fully reused. The adaptive aggregation module combines  feature maps from all the previous blocks to form new feature maps as the input of subsequent blocks. After an adaptive aggregation module or a dense block, a convolution layer is used to compress the feature maps. The aggregated feature maps are upsampled to $H \times W \times C$ ($C$ is the number of the classes for the labels) and the pixel-wise cross entropy loss is computed.}
	\label{fdnet}
\end{figure*}
\section{Fully dense neural networks}
In this section, we introduce the proposed fully dense neural network (FDNet), which is visualized in Fig. \ref{fdnet} comprehensively. We first introduce the whole architecture. Next, the adaptive aggregation structure for dense feature maps is presented in detail. At last, we show the boundary-aware loss function.
\subsection{Encoder-decoder architecture}
Our model (Fig. \ref{fdnet}) is based on the deep encoder-decoder architecture (e.g. \cite{deconvnet,segnet}). The encoder module extracts features from an image and the decoder module produces semantic segmentation prediction.\\
\noindent \textbf{Encoder}.  Our encoder network is based on the DenseNet-264 \cite{densenet_huang} while removing the softmax and fully connected layers of the original network (from the starting convolutional layer to the dense block 4 in Fig. \ref{fdnet}). The input of each convolutional layer within a dense block is the concatenation of all  outputs of its previous layers at a given resolution. Given that $x_{l}$ is the output of the  $\ell^{th}$ layer in a dense block, $x_\ell$ can be computed as follows:
\begin{align}
x_\ell=H_l([x_0,x_1,...,x_{\ell-1}])
\end{align}
where $[x_0,x_1,...,x_{\ell-1}]$ denotes  the concatenation operation of the feature maps $x_0,x_1,...,x_{\ell-1}$, and $x_0$ is the input feature map of the dense block. Meanwhile, $H_\ell$($\cdot$) is defined as a composite function of operations: BN, ReLU, a $1\times 1$ convolution operation followed by BN, ReLU, a $3\times3$ convolution operation. As a result, the output of a dense block includes feature maps from all the layers in this block. Each dense block is followed by a transition layer, which is to compress the number and  size of the feature maps through $1\times 1$ convolution and pooling layers. For an input image $I$, the encoder network produces 4 feature maps $(B_{1}, B_{2}, B_{3}, B_{4})$ with decreasing spatial resolution $(\frac{1}{4}, \frac{1}{8}, \frac{1}{16}, \frac{1}{32})$. In order to reduce spatial information loss, we can remove the pooling layer before the  dense block 4 so that the output feature map of the last dense block (i.e. $B_4$) in the encoder module is  $\frac{1}{16}$ of the size. Atrous convolution is also used to control the spatial density of computed feature responses in the last block as suggested in \cite{deeplabv3}. We refer to this architecture  as FDNet-16s. The original architecture can be taken as FDNet-32s.\\
\noindent \textbf{Decoder}.
As the encoder-decoder structure has much more layers than the original encoder network, how to boost  gradient backpropagation and information flow becomes another problem we have to deal with.  The decoder module progressively enlarges the feature maps while densely reusing previous feature maps by aggregating them into a new feature map. As the input feature map of each dense block has a direct connection to the output of the block, the inputs of  previous blocks in the encoder module are also directly connected to the new feature map . The new feature map is then upsampled to compute loss with the groundtruth, which leads to multiple losses computation. Thus, the inputs of all dense blocks in the FDNet  have a direct connection to the loss layers. This will significantly boost the gradient backpropagation.\\
\indent Following the  DenseNet structure, we also use dense blocks at each stage of the same size after a compression layer with  convolution operation, which will change the number of feature maps from adaptive aggregation structure. The compression layer is composed of BN, ReLU and a $1\times 1$ convolution operation. In the two compression layers after adaptive aggregation, their filter numbers are set to 1024 and 768. In the two compression layers after block 5 and block 6, the filter numbers are  set to 768 and 512. For block 5 and block 6, there are 2 convolutional layers in each of them.
\subsection{Adaptive aggregation of dense feature maps}
\begin{figure}[t!]
	\centering
	\includegraphics[width=0.98\linewidth]{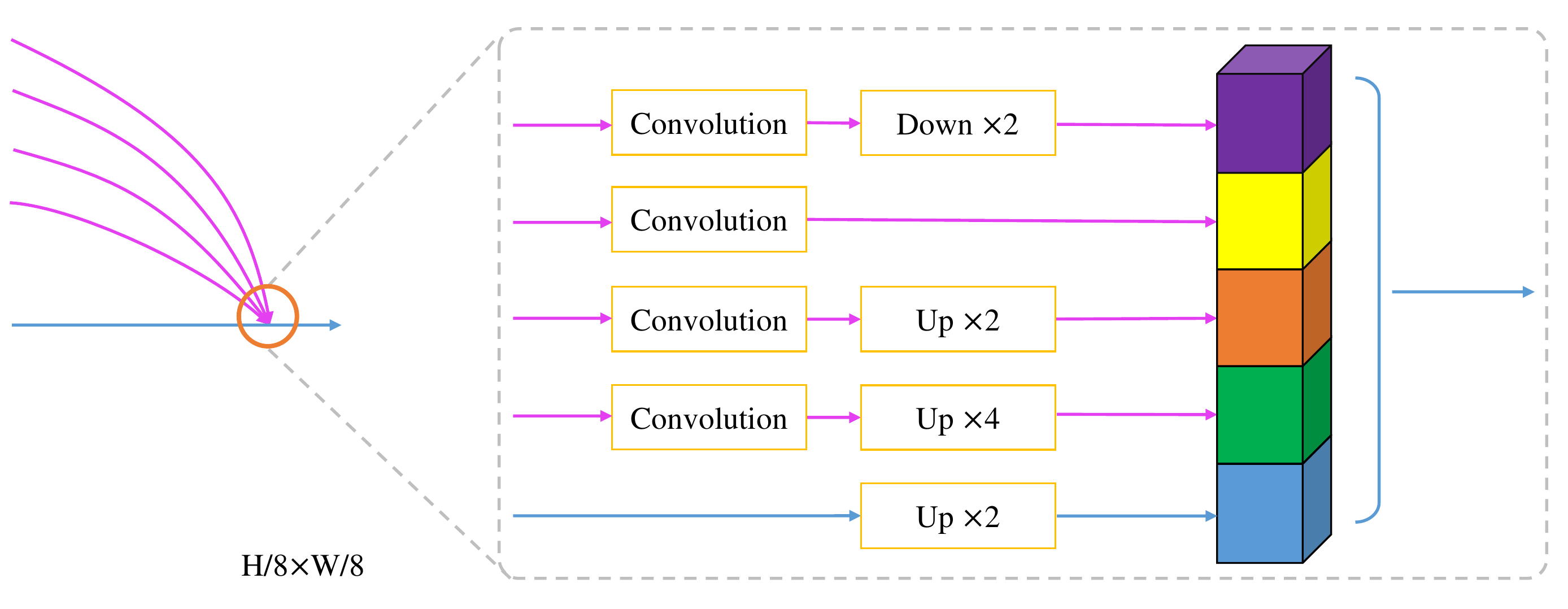}
	\caption{An example of an adaptive aggregation  structure for dense feature maps. For all the input feature maps (not including direct connected input feature map, i.e. blue line), a compression layer with BN, ReLU and $1\times 1$ convolution is applied to adjust the number of feature maps. Then
		an upsampling or downsampling layer is first operated   so that all the feature maps are consistent in size with the output feature map.  They are then concatenated to form a new feature map with $\frac{1}{8}$ of the size of the input image.}
	\label{concat}
\end{figure}

In previous works, e.g. U-Net \cite{u-net} for semantic segmentation, FPN \cite{fpn} for object detection and SharpMask \cite{sharpmask} for instance segmentation, feature maps are reused directly in the corresponding decoder module by concatenating or adding the feature maps. Furthermore, RefineNet \cite{refinenet}, LRR \cite{LRR} and G-FRNet \cite{gateFrNet} refine the feature maps progressively  stage by stage. Instead of just using previous feature maps as before, we introduce an adaptive aggregation structure to make better use of the feature maps from previous blocks. As shown in Fig. \ref{concat}, the feature maps from previous blocks are densely concatenated together by using the  adaptive aggregation structure.\\
\indent The adaptive aggregation structure takes all the feature maps from previous blocks ($B_{1}, B_{2},...$) as input. The feature maps from the lower layers (e.g. $B_{1}, B_{2}$) are of high resolution with coarse semantic information, whereas feature maps from the higher layers (e.g. $B_{3}, B_{4}$) are of low resolution with  rich semantic information. The adaptive aggregation structure combines all previous feature maps to generate rich contextual information and also spatial information. For incoming feature maps, the scale sizes may be different. As shown in Fig. \ref{concat}, the output feature map is  $\frac{1}{8}$ of the size of the input image. To reduce memory consumption, we firstly use the convolutional layer to compress the incoming feature maps except for the direct connected feature map (which has been compressed). The compression layer is also composed of BN, ReLU and a $1\times 1$ convolution operation. In order to make all feature maps consistent in size, we use the convolutional layer to downsample and the deconvolutional layer to upsample the feature maps. Intuitively, we directly concatenate the feature map if it is equivalent to the size of the output feature map. The convolutional layers are all  composed of BN, ReLU and a $3\times 3$ convolution operation with different strides. The deconvolutional layers are all composed of BN, ReLU and a $4 \times 4$  deconvolutional operation with different strides. At last, all the resultant feature maps $D^{i}_{1},D^{i}_{2},...,D^{i}_{M}$ ($M$ input feature maps) are concatenated into a new feature map $F^{i}$ for the $i^{th}$ stage, which is then fed to latter loss computation operation or dense block. The formulation for obtaining the $i^{th}$ dense feature map from the previous feature maps can be written as follows:
\begin{align}
D^{i}_{1}&=T^{i}_{1}(B_{1}),D^{i}_{2}=T^{i}_{2}(B_{2}),...,D^{i}_{M}=T^{i}_M(B_{M}) \notag\\ 
F^{i}&=[D^{i}_1,D^{i}_2,...,D^{i}_{M}]
\end{align}
where $T(\cdot)$ denotes the transformation operation (downsample or upsample). If $B_{j}$ is of the same size as the output feature map, no operation is performaned on $B_{j}$. In addition, $[\cdots]$ stands for  the concatenation operation. \\
\indent In the adaptive aggregation structures for the three stages of the decoder module, the filter numbers in the compression layer for the reused feature map are set to 384, 256 and 128 respectively. The upsampling and downsampling layers will not change the dimension of feature maps.
\subsection{Boundary-aware loss}
In previous works, cross entropy loss function is often used in pipeline, which treat all pixels equally. As shown in  Fig. \ref{boundary}, we can see that the pixels surrounding the boundary are ``hard examples'', which lead to bad prediction. Based on this observation, we construct a boundary-aware loss function, which guides the network to pay more attention on the pixels near the boundary. The loss function is
\begin{align}
loss(L, L^{gt})=-\frac{1}{N}\sum_{j=1}^{K}\sum_{I_i\in S_j}\sum_{c=1}^{C}\alpha_{j}L_{i,c}^{gt}w(L_{i,c})logL_{i,c}
\end{align}
where $L$ is the result of $softmax$ operation on the output feature map and $L^{gt}$ is the groundtruth. The $I_i$ is the $i$-th pixel in the image $I$ and $C$ is number of  categories. We split all the $N$ pixels of image $I$ into several sets $S_j$  based on the distance between the pixels and the boundary so that $I=\{S_1, S_2, ..., S_K\}$. We apply image dilation operation on the boundary with varying kernel size, which refers to as band width shown in Fig. \ref{boundary}, to obtain different set of pixels surrounding the boundary. $\alpha_j$ is the balancing weight and $w(L_{i,c})$ is an attention weight function. Motivated by \cite{focal_loss}, we test two attention weight functions  ($poly$ and $exp$): $w(L_{i,c})=(1-L_{i,c})^{\lambda}$ and $w(L_{i,c})=e^{-\lambda(1-L_{i,c})}$. The $\lambda$ is used to control attention weight. The ablation experiment results are shown in Table \ref{loss}.\\
\indent In order to further boost the gradient backpropagation and information flow, we compute multiple losses for different aggregated feature map $F^{i}$ motivated by \cite{PSPNet,gateFrNet,sdn}. Specifically, $F^{i}$ is fed to upsample module to obtain a feature map $L^{i}$ with channel $C$, where $C$ is number of classes in prediction labels. Then the feature map $L^{i}$ is upsampled by using bilinear interpolation method directly to produce feature map $H\times W\times C$, which is used to compute pixel-wise loss with groundtruth. In terms of formula, the final loss $L_{final}$ is computed as follows:
\begin{align}
L^{i}&=softmax(U_i(F^{i}))\notag\\
L_{final}&=\sum_{i}loss(L^{i}, L^{gt})
\end{align}
where $U_i(\cdot)$ denotes a upsample module with bilinear interpolation operation. \\
\indent In the encoder module, the output feature map of each module is the concatenation of all the feature maps within this block, including the input. And the aggregated feature map is  feature maps from all the previous blocks. Thus, each feature map in the encoder  has much shorter path to loss compared with previous encoder-decoder structure \cite{refinenet,gateFrNet}. The gradient backpropagation and information flowing is much more efficient. This will further boost our network optimization.
\begin{figure}[t!]
	\centering
	\includegraphics[width=0.98\linewidth]{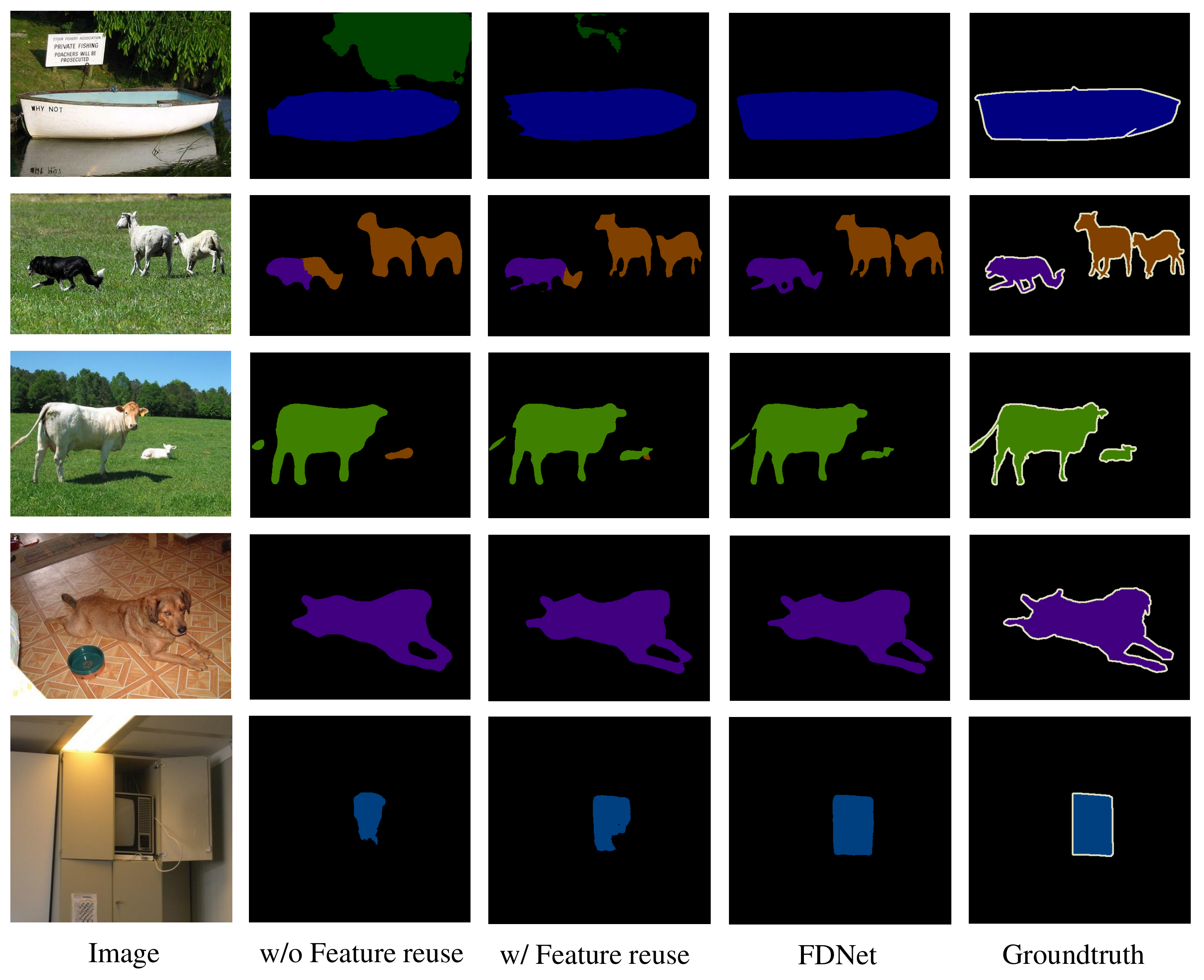}
	\caption{The effect of employing the proposed fully dense feature map reuse structure compared with other frameworks. Our proposed FDNet shows better results (Column \textbf{4}), especially on the \textbf{boundary localization}, compared with the results (Column \textbf{3}) of encoder-decoder structure with feature reuse method (Fig. \ref{encoder} \textbf{Middle})  and the results (Column \textbf{2}) of encoder-decoder structure without feature reuse method (Fig. \ref{encoder} \textbf{Top}). }
	\label{skip}
\end{figure}
\subsection{Implementation details}
\textbf{Training}: The proposed FDNet is implemented with PyTorch on a single NVIDIA GTX 1080Ti. The weights of DenseNet-264 are directly employed in the encoder module of FDNet. In the training step, we adopt data augmentation similar to \cite{deeplab_v2}. Random crops of $512\times512$  and horizontal flip is applied. We train the dataset with 30K iterations. We optimize the network by using the ``poly'' learning rate policy where the initial learning rate is multiplied by $(1-\frac{iter}{max\_iter})^{power}$ with $power=0.9$. The initial learning rate is set to $0.00025$. We set momentum to $0.9$ and weight decay to $0.0005$.\\
\noindent \textbf{Inference}: In the inference step, we pad images with mean value  before feeding full images into the network. We apply multi-scale inference, which is commonly used in semantic segmentation methods \cite{refinenet,sdn}. For multi-scale inference, we average the predictions on the same image across different scales for the final prediction. We set the scales ranging from 0.6 to 1.4. Horizontal flipping is also adopted in the inference. In the ablation experiments, we just use the single scale (i.e. scale = 1.0) and horizontal flipping method to do inference. In addition, we use the last-stage feature map of the decoder module to generate final prediction label map.
\section{Experiments}
In this section,  we describe configurations of experimental datasets  and show  ablation experiments on PASCAL VOC 2012. At last, we report the results on two benchmark datasets: PASCAL VOC 2012 and NYUDv2. 
\begin{table}[t!]
	\caption{The mean IoU scores (\%) for encoder-decoder with different feature map reuse methods on PASCAL VOC 2012 validation dataset. }
	\label{fdnet16s}
	\centering
	\begin{tabular}{c|c|c|c}
		\toprule
		\textbf{Encoder stride}  &\tabincell{c}{w/o feature \\reuse} &\tabincell{c}{w/ feature \\reuse} &\tabincell{c}{dense feature \\reuse}  \\ \hline\hline
		32      		    &77.2 &78.5 &78.9 \\
		16                  &78.2 &79.1 &79.4 \\
		\bottomrule
	\end{tabular}
\end{table}
\begin{table}[t!]
	\caption{ The mean IoU scores (\%) for boundary-aware loss on PASCAL VOC 2012 validation dataset. The $poly$ and $exp$ represent different weighting methods.}
	\label{loss}
	\centering
	\begin{tabular}{c|l|c}
		\toprule
		\textbf{loss}       &       &\textbf{mIoU}  \\ \hline\hline
		CE              &           &79.4 \\
		\hline 
		b-aware($poly$)      &$kernel=(10,20,30,40), \lambda=0$       &\\
		
		&$\alpha=(5,4,3,2,1)$       &79.5\\
		&$\alpha=(8,6,4,2,1)$       &\textbf{80.3}\\
		\hline
		b-aware($poly$)    &$\alpha=(8,6,4,2,1), \lambda=0$       &\\
		&$kernel=(5,10,15,20)$       &79.6\\
		\hline
		b-aware($poly$)    &\tabincell{l}{$\alpha=(8,6,4,2,1)$,\\$kernel=(10,20,30,40)$ }      &\\
		&$ \lambda=1$       &80.0\\		
		&$ \lambda=2$       &79.6\\
		&$ \lambda=5$       &77.7\\  
		\hline		
		b-aware($exp$)    &\tabincell{l}{$\alpha=(8,6,4,2,1)$,\\$kernel=(10,20,30,40)$ }      &\\
		&$ \lambda=0.25$       &80.7\\		
		&$ \lambda=0.5$       &80.3\\
		&$ \lambda=0.75$       &\textbf{80.9}\\  
		&$ \lambda=1$       &80.6\\		
		&$ \lambda=2$       &79.2\\
		
		\bottomrule
	\end{tabular}
\end{table}
\subsection{Datasets description}
To show the effectiveness of our approach, we conduct comprehensive experiments on PASCAL VOC 2012 dataset \cite{pascal_voc} and NYUDv2 dataset \cite{nyudv2}. \\
\indent PASCAL VOC 2012: The dataset has 1,464 images for training, 1,449 images for validation and 1,456 images for testing, which involves 20 foreground object classes and one background class. Meanwhile, we augment the training set with extra labeled PASCAL VOC images provided by Semantic Boundaries Dataset \cite{sem_boudary}, resulting in 10,582 images as \emph{trainaug} dataset for training. \\
\indent NYUDv2: The NYUDv2 dataset \cite{nyudv2} consists of 1449 RGB-D images showing indoor scenes. We use the segmentation labels provided in \cite{nyudv2_generate}, in which all labels are mapped to 40 classes. We use the standard training/test split with 795 and 654 images, respectively. Only RGB images are used in our experiments.\\
\indent Moreover, we perform a series of ablation evaluations on PASCAL VOC 2012 dataset with mean IoU score reported. We use the \emph{trainaug}  and validation dataset of PASCAL VOC 2012 for  training and inference,  respectively. 
\begin{table}[t!]
	\caption{GPU memory, number of parameters and some results on VOC 2012 test dataset are reported.}
	\label{memory}
	\centering
	\begin{tabular}{l|c|c|c}
		\toprule
		\textbf{Methods}      &RefineNet-152　　&  FDNet          &$\text{SDN}_{\text{M2}}$     \\ \hline\hline
		GPU Memory (MB)          &4253        &  \textbf{2907}   & - \\
		Parameters (M)        &109.2          &  \textbf{113.1} &161.7 \\
		mIOU                  &83.4           &  \textbf{84.2}  &83.5 \\
		\bottomrule
	\end{tabular}
\end{table}
\begin{table}
	\caption{Comparison of different mothods on PASCAL VOC 2012 validation dataset with mean IoU score (\%). \emph{FDNet-16s-MS} denotes the evaluation on multiple scales. \emph{FDNet-16s-finetuning-MS} denotes fine-tuning on standard training data (1464 images) of PASCAL VOC 2012 dataset after training on the \emph{trainaug} dataset.}
	\label{voc_val}
	\centering
	\begin{tabular}{l|c}
		\toprule
		\textbf{Method}   &\textbf{mIoU}\\ \hline\hline
		Deeplab-MSc-CRF-LargeFOV       		    &68.7 \\
		DeconvNet                          &67.1 \\
		DeepLabv2                         &77.7 \\
		G-FRNet                             &77.8 \\
		DeepLabv3                         &79.8 \\ 
		SDN                                     &80.7 \\
		DeepLabv3+                       &81.4 \\
		\hline
		FDNet-16s                                                &80.9  \\
		FDNet-16s-MS                                            &82.1 \\
		FDNet-16s-finetuning-MS                                    &\textbf{84.1}\\
		\bottomrule
	\end{tabular}
\end{table}
\subsection{Feature map reuse}
\begin{figure}[t!]
	\centering
	\includegraphics[width=0.99\linewidth]{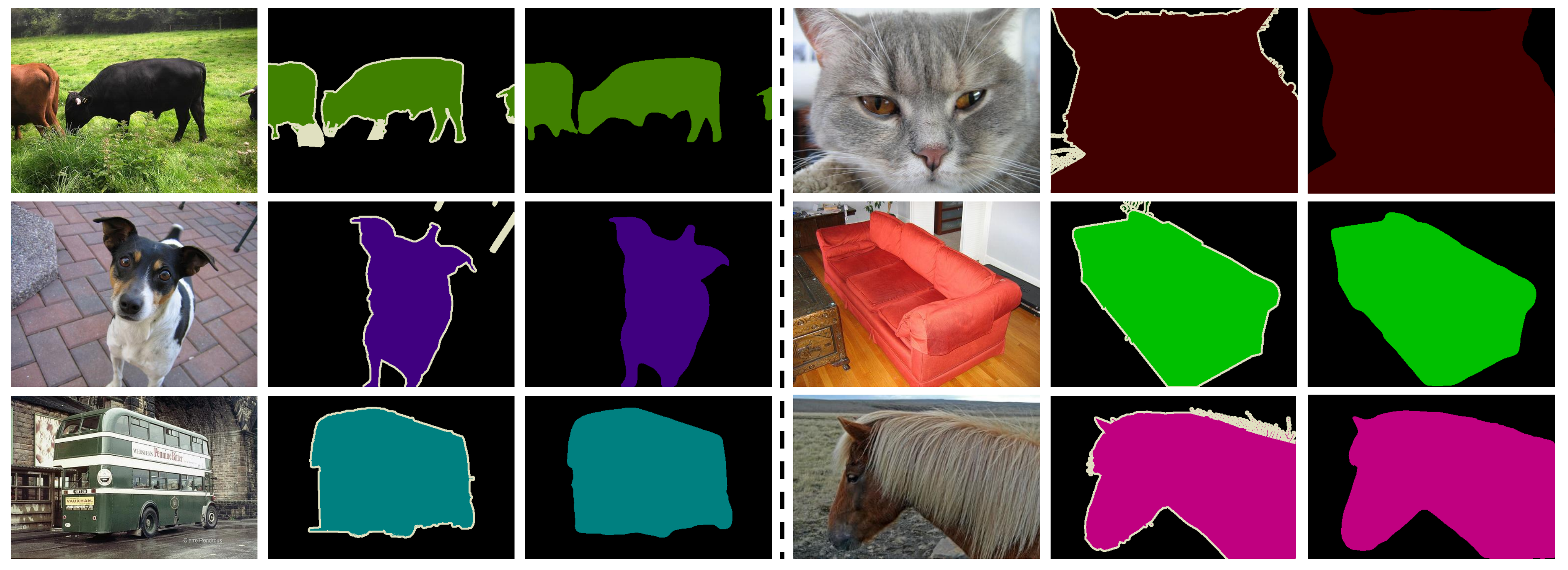}
	\caption{Some visual results on PASACAL VOC 2012 dataset. Three columns of each group are image, groundtruth and prediction label map.}
	\label{voc_vis}
\end{figure}
\begin{table*}[t!]
	\caption{Quantitative results (\%) in terms of mean IoU on PASCAL VOC 2012 test set. Only VOC data is used as training data and denseCRF \cite{densecrf} is not included.}
	\label{voc2012}
	\resizebox{\textwidth}{!}{ %
		\centering
		\begin{tabular}{l|c|c|c|c|c|c|c|c|c|c|c|c|c|c|c|c|c|c|c|c|c}
			\toprule
			\textbf{Method} &\rotatebox{90}{aero} &\rotatebox{90}{bicycle} &\rotatebox{90}{bird} &\rotatebox{90}{boat} &\rotatebox{90}{bottle} &\rotatebox{90}{bus} &\rotatebox{90}{car} &\rotatebox{90}{cat}  &\rotatebox{90}{chair} &\rotatebox{90}{cow} &\rotatebox{90}{table} &\rotatebox{90}{dog} &\rotatebox{90}{horse} &\rotatebox{90}{mbike} &\rotatebox{90}{person} &\rotatebox{90}{plant} &\rotatebox{90}{sheep} &\rotatebox{90}{sofa} &\rotatebox{90}{train} &\rotatebox{90}{tv} &\textbf{mIoU} \\ \hline\hline
			
			
			DeconvNet \cite{deconvnet} &89.9 &39.3 &79.7 &63.9 &68.2 &87.4 &81.2 &86.1 &28.5 &77.0 &62.0 &79.0 &80.3 &83.6 &80.2 &58.8 &83.4 &54.3 &80.7 &65.0 &72.5 \\
			Deeplabv2 \cite{deeplab_v2}  &84.4 &54.5 &81.5 &63.6 &65.9 &85.1 &79.1 &83.4 &30.7 &74.1 &59.8 &79.0 &76.1 &83.2 &80.8 &59.7 &82.2 &50.4 &73.1 &63.7 &71.6\\
			GCRF \cite{GCRF}       &85.2 &43.9 &83.3 &65.2 &68.3 &89.0 &82.7 &85.3 &31.1 &79.5 &63.3 &80.5 &79.3 &85.5 &81.0 &60.5 &85.5 &52.0 &77.3 &65.1 &73.2 \\
			Adelaide \cite{piecewise}&90.6 &37.6 &80.0 &67.8 &74.4 &92.0 &85.2 &86.2 &39.1 &81.2 &58.9 &83.8 &83.9 &84.3 &84.8 &62.1 &83.2 &58.2 &80.8 &72.3 &75.3\\
			LRR \cite{LRR}          &91.8 &41.0 &83.0 &62.3 &74.3 &93.0 &86.8 &88.7 &36.6 &81.8 &63.4 &84.7 &85.9 &85.1 &83.1 &62.0 &84.6 &55.6 &84.9 &70.0 &75.9\\
			G-FRNet \cite{gateFrNet} &91.4 &44.6 &91.4 &69.2 &78.2 &95.4 &88.9 &93.3 &37.0 &89.7 &61.4 &90.0 &91.4 &87.9 &87.2 &63.8 &89.4 &59.9 &87.0 &74.1 &79.3 \\
			PSPNet \cite{PSPNet}   &91.8 &71.9 &\textbf{94.7} &71.2 &75.8 &95.2 &89.9 &95.9 &39.3 &90.7 &71.7 &90.5 &94.5 &88.8 &89.6 &\textbf{72.8} &89.6 &64.0 &85.1 &76.3 &82.6 \\ 
			SDN \cite{sdn}        &\textbf{96.2} &73.9 &94.0 &74.1 &76.1 &\textbf{96.7} &89.9 &\textbf{96.2} &\textbf{44.1} &92.6 &\textbf{72.3} &91.2 &94.1 &89.2 &\textbf{89.7} &71.2 &93.0 &59.0 &\textbf{88.4} &76.5 &83.5\\
			\hline
			FDNet                 &95.5 &\textbf{79.9} &88.6 &\textbf{76.1} &\textbf{79.5} &\textbf{96.7} &\textbf{91.4} &95.6 &40.1 &\textbf{93.0} &71.5 &\textbf{93.4} &\textbf{95.7} &\textbf{91.1} &89.2 &69.4 &\textbf{93.3} &\textbf{68.0} &88.3 &\textbf{76.8} &\textbf{84.2}\\
			\bottomrule
		\end{tabular}
	}
\end{table*}
To verify the power of dense feature maps reuse, we compare our method with other two baseline frameworks. In this experiment, cross entropy loss is used. One is encoder-decoder structure without feature map reuse (Fig. \ref{encoder} Top) and the other is encoder-decoder structure with naive feature map reuse (Fig. \ref{encoder} Middle). We also compare the three frameworks on different encoder strides (the ratio of input image resolution to smallest output feature map of encoder, i.e. 16 and 32). \\
\indent The results are shown in Table \ref{fdnet16s}. It is observed that the performance  increases when feature maps are reused. Specifically, the performance for encoder-decoder (encoder stride = 32) without feature map reuse is only 77.2\%. After the naive feature map reuse, the performance can increase to 78.5\%. Furthermore, our fully dense feature map reuse can further improve the performance to 78.9\%.  In addition, when we adopt the stride 16 for the encoder module, the performance is much better than the original encoder with stride 32 on the three frameworks. This is because the spatial information loss is reduced by the encoder with smaller stride. We speculate that encoder with stride 8 can have better result similar to \cite{deeplabv3,deeplabv3+}. Because of memory limitation, we only test on the encoder with stride 16 and 32. \\
\indent We also show some predicted semantic label maps for different feature map reuse methods in Fig. \ref{skip}. For the encoder-decoder structure without feature map reuse, the result is poor, especially for boundary localization. Though the naive feature map reuse method improves the segmentation result partially, it is still hard to obtain accurate pixel-wise prediction. The fully dense feature map reuse method shows very excellent results on the boundary localization.
\subsection{Boundary-aware loss}
In order to demonstrate the effect of proposed boundary-aware loss method, we take FDNet-16s as baseline to test the performance of different parameters.
 We mainly use the $kernel=(10,20,30,40)$ and $kernel=(5,10,15,20)$ by splitting  the pixels into $K=5$ sets (the remaining pixels are referred to as $S_5$). For $poly$ weight method, the boundary-aware loss method (b-aware) degrades into cross entropy method (CE) when $\alpha=(1,1,1,1,1)$ and $\lambda=0$. As shown in Table \ref{loss}, the simply weighting on the pixels surrounding the boundary shows better performance compared with general cross entropy method, which enhances the performance by $0.9\%$. By fixing the $\alpha$ and $kernel$, we try different parameter $\lambda$ in Table \ref{loss}. Comparing  the $poly$ and $exp$ methods, we can observe that $exp$ brings obvious improvement by $1.5\%$. On the contrary, the $poly$ methods lead to worse effect compared with baseline method ($80.0$ vs $80.3$). In addition, the network cannot converge  for $\lambda<1$. We also compare the labeling accuracy for the pixels near the boundary. As shown in Fig. \ref{boundary}, the FDNet with boundary-aware loss shows obvious better performance for the pixels surrounding the boundary.
\subsection{Memory analysis}
For semantic segmentation task, memory consumption and parameter number are both important issues. The proposed FDNet uses fully dense connected structure with nearly the same number of parameters compared with RefineNet \cite{refinenet}. As shown in Table \ref{memory}, the FDNet consumes much less GPU memory (training process) compared with RefineNet. In addition, the memory consumption of FDNet can be reduced by using sharing memory efficiently based on \cite{densenet_memory}. Compared with SDN \cite{sdn}, there are much less parameters for FDNet but the performance is much better.
\subsection{PASCAL VOC 2012}
We evaluate the performance on the PASCAL VOC 2012 dataset following previous works \cite{refinenet,PSPNet}. As FDNet-16s shows a better performance (Table \ref{fdnet16s}), we only report the performance of FDNet-16s in the following experiments. We adopt the  boundary-aware method in the training step. As shown in Table \ref{voc_val}, FDNet-16s achieves very comparable result with an 82.1\% mean IoU accuracy compared with previous works (\cite{deeplabv3,gateFrNet,sdn}) when evaluated on multiple scales. Moreover, after  fine-tuning the model  on the standard training data (1464 images) of the PASCAL VOC 2012 dataset, we achieve a much better result with 84.1\% mean IoU accuracy, which is the best result currently if not considering pretraining on other dataset (such as MS-COCO \cite{mscoco}). Some visual results with image, groundtruth and prediction label maps are shown in Fig. \ref{voc_vis}. \\
\indent Table \ref{voc2012} shows the quantitative results of our method on the test dataset, where we only report the results using the PASCAL VOC dataset. We achieve the best result of 84.2\% on test data without pretraining on other datasets, which is the highest score  when  considering training on  PASCAL VOC 2012 dataset.  Though latest work DeepLabv3+ \cite{deeplabv3+}  achieves a mean IoU score of 89.0\% on test data of PASCAL VOC 2012, the result relies upon pretraining on a much larger dataset MS-COCO \cite{mscoco} or JFT \cite{jft}. In fact, FDNet-16s shows very comparable result compared with DeepLabv3+ on the validation dataset (Table \ref{voc_val}).
\subsection{NYUDv2 Dataset}
We conduct experiments on the NYUDv2 dataset to compare FDNet-16s with previous works. We follow the training setup in PASCAL VOC 2012 and multi-scale inference is also adopted. The results are reported in Table \ref{nyud2}. Similar to \cite{refinenet}, pixel accuracy, mean accuracy and mean IoU are used to evaluate all the methods. Some works make use of both depth image and RGB image as input and obtain very better results. For example,  RDF \cite{rdfnet} achieves 50.1\% (mean IoU) by using depth information. For a fair comparison, we only report the results  training on only RGB images. As is shown, FDNet-16s outperforms previous work in terms of all  metrics. In particular, our result is better than RefineNet \cite{refinenet} by 0.9\% in terms of mean IoU accuracy. 
\begin{table}
	\caption{Quantitative results (\%) on NYUDv2 dataset (40 classes). The model is only trained on the provided training image dataset. }
	\label{nyud2}
	\centering
	\begin{tabular}{l|c|c|c}
		\toprule
		\textbf{Method}  &pixel acc. &mean acc. &\textbf{mIoU}  \\ \hline\hline
		
		SegNet                    &66.1 &36.0 &23.6 \\
		
		Bayesian SegNet    &68.0 &45.8 &32.4 \\
		FCN-HHA        		   &65.4 &46.1 &34.0 \\
		Piecewise    		   &70.0 &53.6 &40.6 \\	
		RefineNet  		   &73.6 &58.9 &46.5 \\
		\hline
		FDNet-16s                           	  &\textbf{73.9} &\textbf{60.3} &\textbf{47.4}\\
		\bottomrule
	\end{tabular}
\end{table}
\section{Conclusion}
In this paper, we have presented the fully dense neural network (FDNet) with an encoder-decoder structure for semantic segmentation.  For each layer of the FDNet in the decoder module, feature maps of almost all the previous layers are aggregated as the input. Furthermore, we propose a boundary-aware loss function by paying more attention to the pixels surrounding the boundary. The proposed FDNet is very advantageous to semantic segmentation. On the one hand, the class boundaries as  spatial information are well reconstructed by using the Encoder-Decoder structure with a boundary-aware loss function. On the other hand, the FDNet learns more efficiently  with the more efficient  gradient backpropagation, much  similar to the arguments already demonstrated in ResNet and DenseNet. The  experiments show that our model outperforms  previous works on two public benchmarks when  training on other datasets is not considered. \\
\section{Acknowledgments} 
This work is supported by GRF 16203518, Hong Kong RGC 16208614,  T22-603/15N, Hong Kong ITC PSKL12EG02, and China 973 program, 2012CB316300.

%
\bibliography{AAAI-ZhenM.4047}
\bibliographystyle{aaai}

\end{document}